\documentclass[10pt,twocolumn,letterpaper]{article}

\usepackage[pagenumbers]{cvpr} 

\definecolor{cvprblue}{rgb}{0.21,0.49,0.74}
\usepackage[pagebackref,breaklinks,colorlinks,allcolors=cvprblue]{hyperref}
\usepackage{svg}
\usepackage{amsmath}
\usepackage{xspace} 
\usepackage{pifont}
\usepackage{subcaption}
\usepackage{adjustbox}
\usepackage{colortbl}
\usepackage{algorithm}
\usepackage{algpseudocode}
\usepackage{multirow}
\usepackage{xcolor}
\usepackage{units}
\usepackage{amsfonts,amssymb}
\usepackage{caption}

\def\paperID{8384} 
\def\confName{CVPR}
\def\confYear{2025}

\title{Token Pruning for Caching Better: 9$\times$  Acceleration on Stable Diffusion for Free}

\def\spaces{~~~~~}
\author{Evelyn Zhang$^{1}$\thanks{Equal contribution}, \spaces{}  
Bang Xiao$^1$\footnotemark[1], \spaces{} 
Jiayi Tang$^2$, \spaces{}
Qianli Ma$^1$,\spaces{}
Chang Zou$^{3,1}$, \\ 
Xuefei Ning$^4$, \spaces{}
Xuming Hu$^5$, \spaces{}
Linfeng Zhang$^1$\thanks{Corresponding Author: \texttt{zhanglinfeng@sjtu.edu.cn}}  \\\\
$^1$Shanghai Jiao Tong University ~~
$^2$China University of Mining and Technology\\ 
$^3$University of Electronic Science and Technology of China ~~
$^4$Tsinghua University  \\
$^5$The Hong Kong University of Science and Technology (Guangzhou)
}

\begin{document}
 \maketitle
 \begin{abstract}
Stable Diffusion has achieved remarkable success in the field of text-to-image generation, with its powerful generative capabilities and diverse generation results making a lasting impact.
However, its iterative denoising introduces high computational costs and slows generation speed, limiting broader adoption.
The community has made numerous efforts to reduce this computational burden, with methods like feature caching attracting attention due to their effectiveness and simplicity. Nonetheless, simply reusing features computed at previous timesteps causes the features across adjacent timesteps to become similar, reducing the dynamics of features over time and ultimately compromising the quality of generated images.
In this paper, we introduce a dynamics-aware token pruning (DaTo) approach that addresses the limitations of feature caching. DaTo selectively prunes tokens with lower dynamics, allowing only high-dynamic tokens to participate in self-attention layers, thereby extending feature dynamics across timesteps. DaTo combines feature caching with token pruning in a training-free manner, achieving both temporal and token-wise information reuse. Applied to  Stable Diffusion on the ImageNet, our approach delivered a 9$\times$ speedup while reducing FID by 0.33, indicating enhanced image quality. On the COCO-30k, we observed a 7$\times$ acceleration coupled with a notable FID reduction of 2.17.
\\ 
\textbf{Code: \href{https://github.com/EvelynZhang-epiclab/DaTo}{\texttt{\textcolor{cyan}{github.com/EvelynZhang-epiclab/DaTo}}}}
\end{abstract}
 \section{Introduction}
\label{sec:intro}
\begin{figure}[t]
\centering
\includegraphics[width=\linewidth]{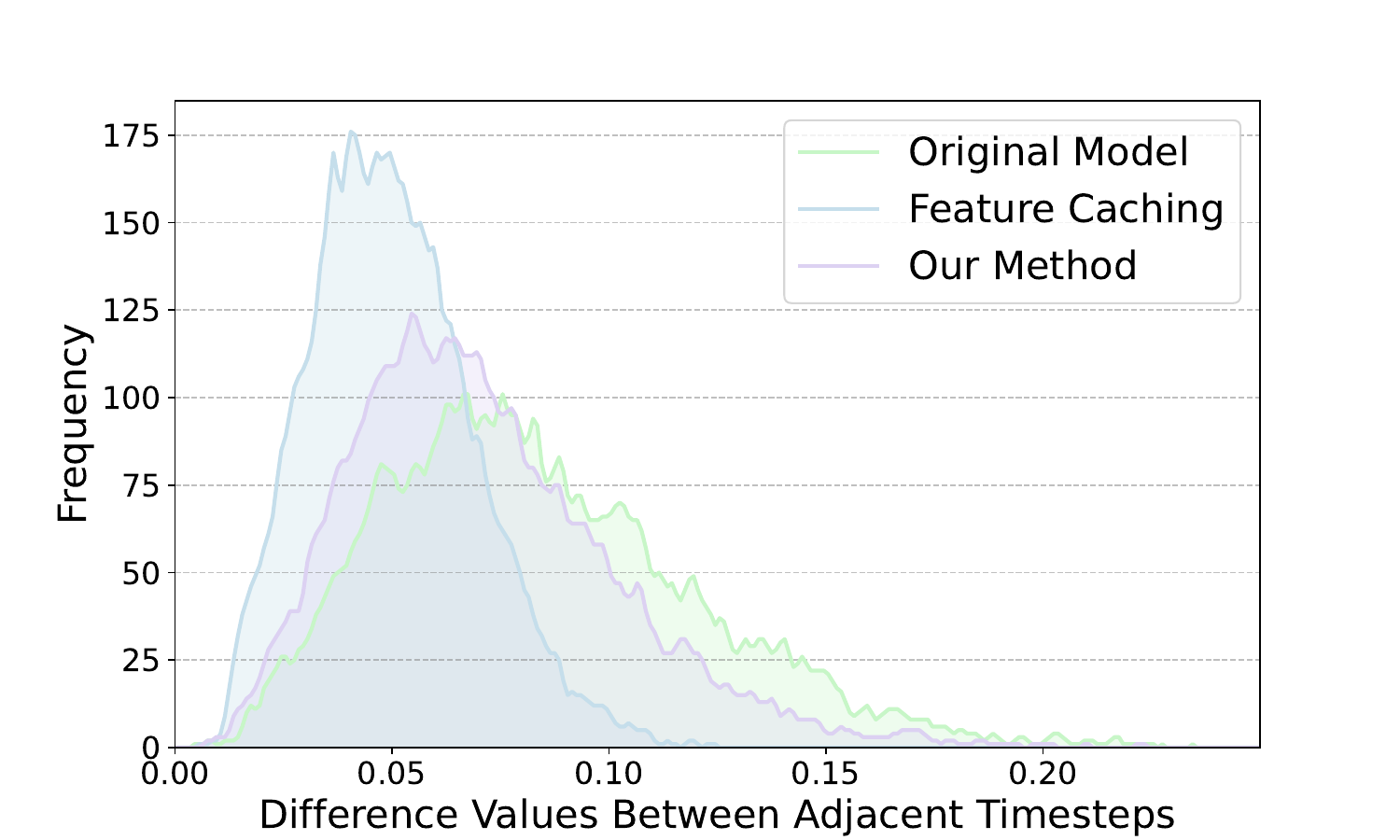}
\vspace{-0.6cm}
\caption{
Feature difference between adjacent timesteps under three acceleration methods: original Stable Diffusion, feature caching~\cite{sd-deepcache}, and DaTo. DaTo produces a distribution that is more similar to the original Stable Diffusion, suggesting that our proposed token pruning method helps restore feature dynamics across timesteps.
}
\vspace{-0.3cm}
\label{fig:cache_stats}
\end{figure}
\begin{figure*}[t]
\centering
\small
\includegraphics[width=1\textwidth]{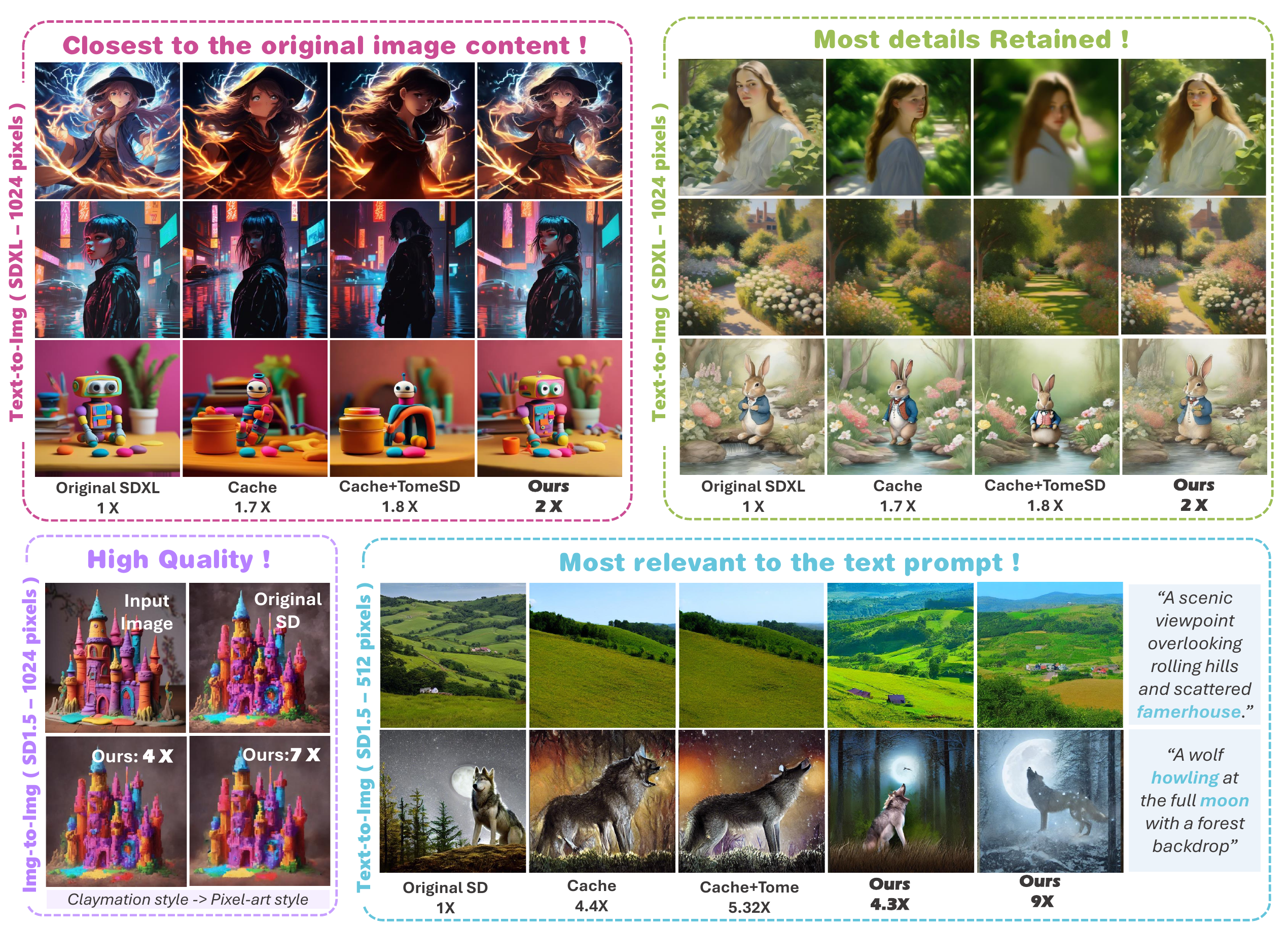}
\vspace{-20pt}
\caption{Visual comparison across multiple models and methods, including SD and SDXL. The methods include caching only (DeepCache), token pruning only (ToMeSD), and a naive combination of both (DeepCache \& ToMeSD). Our proposed model maintains high fidelity to the original image content while preserving intricate details and aligning closely with textual prompts, achieving superior image quality in both text-to-image and image-to-image generation tasks.}
\vspace{-12pt}
\label{fig:overall_visualization}
\end{figure*}



Diffusion models~\cite{dm-ddpm,stable-diffusion} have made significant progress in the field of generative modeling in recent years, widely applied to various tasks such as text-to-image generation~\cite{stable-diffusion}, video generation~\cite{video-1,video-2} and text generation~\cite{li2022diffusion}, attracting attention due to their powerful generative capabilities. 
Although diffusion models possess strong generative capabilities, their iterative denoising mechanism leads to substantial computational costs and slow generation speed. To accelerate the reverse process in diffusion models, efforts have been made in some directions: reducing the number of sampling steps~\cite{ddim,plms,PNM,CM} and minimizing the computational cost at each step, such as knowledge distillation~\cite{meng2023distillation,pd}, structural pruning~\cite{diffusion_prune1}, quantization~\cite{diffusion_quant1,diffusion_quant2,diffusion_quant3}, token pruning~\cite{sd-tome} and feature caching~\cite{sd-deepcache,l2cache}.

Among them, feature caching~\cite{sd-deepcache,l2cache}, which stores the features computed in the previous timesteps and then reuses them in the next timesteps for acceleration, has gained abundant popularity thanks to its effectiveness and simplicity. 
However, the feature reuse forces the features in different timesteps to have similar values, reducing the dynamics of features along the timesteps, harming the original diffusion process and thus reducing the generation quality.
Figure~\ref{fig:cache_stats} shows the distribution of the feature differences between adjacent timesteps on each token (pixel). 
It is observed that compared with the original stable diffusion, stable diffusion with feature caching exhibits a significantly lower value, confirming the above conjecture. 
This observation raises the following question - \emph{Is it possible to perform feature caching but still maintain the correct feature dynamics across the timesteps?}

\noindent \textbf{Token Pruning for Better Caching: } Fortunately, as shown in Figure~\ref{fig:cache_stats}, these still exist a few tokens that exhibit clear differences in different timesteps, making it possible to recover the dynamics of features with them.
Based on this observation, we propose to extend the dynamics from these tokens to all the tokens by introducing a dynamics-aware token pruning (DaTo). DaTo is designed to prune the tokens that have smaller dynamics in different timesteps, feed the left tokens to the self-attention layers in stable diffusion, and then recover the pruned tokens with tokens that have the largest dynamics. DaTo brings benefits in three folds: (1) The tokens with smaller dynamics are pruned and then recovered with the tokens with the highest dynamics, which implies extending the dynamics from the few tokens to all the tokens. (2) In self-attention layers, only the tokens with high dynamics attend to other tokens, which avoids the negative influence of the tokens with smaller dynamics. (3) Since the number of tokens in self-attention layers is reduced, token pruning also reduces the computation and memory costs. As shown in Figure~\ref{fig:cache_stats}, DaTo successfully increases the feature dynamics in all the tokens, recovering the distribution of feature dynamics broken by feature caching to the original stable diffusion, allowing the accelerated model to maintain its generation quality.


\noindent \textbf{Searching the Optimal Strategy:} The diffusion model in different timesteps tends to generate different contents and exhibit different redundancy. 
As a result, applying the identical pruning ratio and caching strategy across all timesteps results in limited benefits, as it ignores such differences.
Previous works usually address this problem by manually adjusting the hyper-parameters in different timesteps, which is still not optimal and fails in generalization.
To solve this problem, we propose to formulate this problem as an optimization problem and solve it with evolution methods. Specially, we first define the search space of DaTo and prune the search space to reduce the search costs. Then, we integrate the NSGA-II~\cite{nsga2} evolutionary search algorithm, which optimizes multiple objectives, including inference latency and generation quality.
Such a searching solution allows us to obtain the optimal strategy for pruning and caching with acceptable costs ($\leq$20 GPU hours). Besides, the obtained strategy exhibits better generalization. For instance, the strategy searched from Stable Diffusion on ImageNet works well in SDXL on MSCOCO.


Our method achieves substantial improvements in efficiency and quality across multiple benchmark datasets, \emph{in a training-free manner}. Applied to the Stable Diffusion v1.5 model on the ImageNet~\cite{imagenet}, our approach delivered a 9$\times$ speedup while reducing FID by 0.33, indicating enhanced image quality. On the COCO-30k~\cite{coco}, we observed a 7$\times$ acceleration coupled with a notable FID reduction of 2.17.

Our contributions can be summarized as follows:
\begin{itemize}
    \item We propose dynamics-aware token pruning (DaTo), which avoids the generation quality drop caused by feature caching via pruning the tokens whose dynamics in different timesteps have been reduced by feature caching and recovers them by the tokens with large dynamics.
    \item We propose to search for the optimal feature caching and token pruning strategy via evolutionary methods, which fully unleashes the potential of DaTo.
    \item Extensive experiments on both Stable Diffusion and SDXL demonstrate the effectiveness of our method without requirements for training and additional data. Up to 9$\times$ acceleration can be obtained on Stable Diffusion with no drop in generation quality.
\end{itemize}



 \section{Related Work}
\label{sec:related_work}
\subsection{Efficient Diffusion Models}

Diffusion models~\cite{dm-1,dm-2,dm-3,dm-ddpm} generate images by iteratively denoising a noisy input through multiple diffusion steps. Modern large-scale diffusion models, such as the latent diffusion model~\cite{stable-diffusion}, typically employ a U-Net~\cite{unet} architecture augmented with transformer-based blocks. Recent efforts to enhance the efficiency of diffusion models focus on reducing the number of sampling steps and compressing the denoising networks. Distillation methods, like those in~\cite{DistGDM}, transfer knowledge from multi-step teacher models to student models capable of sampling in fewer steps, even down to a single step. Additionally, fast samplers such as DDIM~\cite{ddim} reduce the number of sampling steps without the need for retraining. Techniques like quantization~\cite{diffusion_quant1,diffusion_quant2}, pruning~\cite{diffusion_prune2}, and distillation~\cite{OAC} have been applied to compress the U-Net. However, most of these methods, except for fast samplers, require retraining the models, which is time-consuming, highlighting the need for more efficient approaches that avoid retraining.

\subsection{Token Reduction}
Token reduction strategies, first introduced for ViTs, enhance efficiency by reducing the quadratic computational burden of self-attention, with both learned and heuristic methods explored. Learned approaches, such as DynamicViT~\cite{Dynavit} and A-ViT~\cite{A-vit}, employ auxiliary models to rank and prune redundant tokens, leveraging techniques like MLPs for pruning masks or feature channels with auxiliary losses. Heuristic methods, including Token Pooling~\cite{TokenPool}, offer practical solutions without extensive training. Training-free methods like Token Merging~\cite{vit-tome} introduce efficient algorithms to merge similar tokens based on bipartite matching, while Token Fusion~\cite{vit-tofo} combines the benefits of merging and pruning by dividing tokens into groups for pruning or merging based on their correlations. In the context of diffusion models, token reduction is less explored. Recent works like ToMeSD~\cite{sd-tome} adapts traditional token reduction techniques to diffusion models but may overlook the unique aspects of generative tasks. AT-EDM~\cite{sd-adt} uses a graph algorithm to identify redundant tokens, but issues like unpredictable convergence and lack of batch computation hinder its practicality.

\subsection{Cache Mechanism}

Cache-based acceleration methods primarily focus on feature reuse across timesteps to improve the efficiency of diffusion models. For U-Net-based models, DeepCache~\cite{DeepCache} leverages the temporal redundancy in the sequential denoising steps by caching and reusing features across adjacent stages, particularly reusing high-level features while updating low-level ones efficiently. Faster Diffusion~\cite{FasterDiffusion} accelerates sampling by caching encoder features and skipping computations at certain timesteps. However, these methods are specific to U-Net architectures and are not directly applicable to Diffusion Transformers (DiT). To address this limitation, methods like $\Delta$-DiT~\cite{DeltaDiT} introduce cache mechanisms tailored for DiT architectures. By analyzing the correlation between DiT blocks and image generation, they find that early blocks are associated with image outlines while later blocks relate to details. Utilizing this insight, $\Delta$-DiT employs a cache mechanism called $\Delta$-Cache to accelerate inference by selectively caching features. Although these cache-based methods offer significant speedups without retraining, optimizing cache strategies remains challenging, particularly in balancing efficiency gains with the preservation of generation quality.
 \section{Methodology}
\label{sec:methodology}
\begin{figure*}[t]
\centering
\small
\includegraphics[width=1\textwidth]{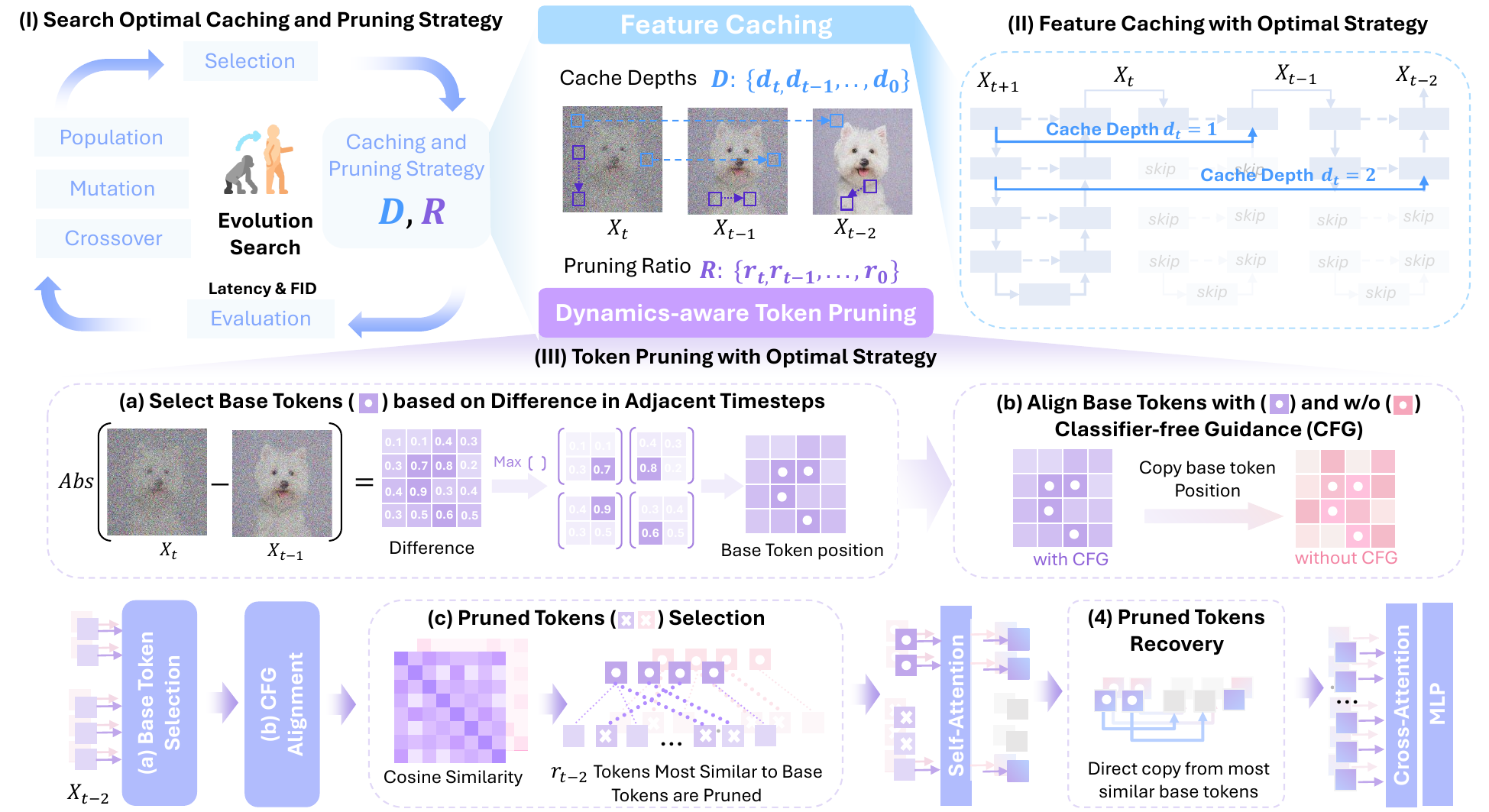}
\vspace{-8pt}
\caption{Overall pipeline of DaTo. \textbf{(I) Search Optimal Caching and Pruning Strategy.} We use evolutionary search to identify the optimal caching depth $d$ and pruning ratio $r$ for each timestep by minimizing both the time latency and FID Score\cite{fid}. \textbf{(II) Feature Caching with Optimal Strategy.} Feature caching employs dynamic token pruning based on the optimal strategy for efficiency. \textbf{(III) Feature Pruning with Optimal Strategy. } \textit{(a):Base Token selection based on adjacent timestep differences: }Divide the image into $s\times s$ patches and select the base token with the largest noise difference between adjacent time steps in each patch. \textit{(b) Align Base Tokens with and w/o CFG guidance: } Make the positions of the base tokens without CFG guidance match the positions of the base tokens with CFG guidance. \textit{(c) Pruned token selection: }$r$ tokens that exhibit the highest consine similarity to the base tokens are chosen as the pruned tokens, and \textit{(d) Pruned Token recovery:} The pruned tokens are restored by copying from the base tokens that are most similar to them.}
\vspace{-15pt}
\label{fig:overall}
\end{figure*}

\subsection{Preliminary}
\subsubsection{Diffusion Models}
A diffusion model consists of a diffusion process that introduces Gaussian noise to a real image, followed by a denoising process that iteratively refines the image from standard Gaussian noise back to the original.
Suppose the data distribution is denoted as $q(x_0)$, where $x_0$ represents the original data. For a given training data sample $x_0 \sim q(x_0)$, the forward diffusion process aims to generate a sequence of progressively noisier latents $x_1, x_2, \ldots, x_T$ through a Markov process described as follows:
\begin{equation}
   q(x_t \mid x_{t-1}) = \mathcal{N}(x_t ; \sqrt{1 - \beta_t} \, x_{t-1}, \, \beta_t \, \mathbf{I}), \forall t \in T, 
\end{equation}

where $t$ denotes the diffusion process step, $T$ is the set of all steps, $\beta_t \in (0,1)$ indicates the difference at each diffusion step, $\mathbf{I}$ is the identity matrix with the same dimensions as the input $x_0$, and $\mathcal{N}(x; \mu, \sigma)$ represents a normal distribution with mean $\mu$ and covariance $\sigma$.

In order to generate a new data sample, diffusion models first sample $x_T$ from a standard normal distribution and subsequently reduce noise by reversing the diffusion process using the intractable distribution $q(x_{t-1} \mid x_t)$. To approximate this reverse distribution, diffusion models traditionally employ a denoising network $p_\theta$, which is parameterized to model the distribution as follows:
\begin{equation}
    p_\theta(x_{t-1} \mid x_t) = \mathcal{N}(x_{t-1}; \mu_\theta(x_t, t), \Sigma_\theta(x_t, t)),
\end{equation}
where $\mu_\theta$ and $\Sigma_\theta$ denote trainable functions for the mean and covariance, respectively.

The UNet-style denoising network $p_\theta$, which comprises $L$ down/up blocks and a single mid block, is denoted as $p_\theta = \left( \prod_{l=0}^{L-1} f^{\text{down}}_{l} \right) \circ f^{\text{mid}}_{L} \circ \left( \prod_{l=L-1}^{0} f^{\text{up}}_{l} \right)$, where $\prod$ denotes a sequential composition of functions. As noted in prior work~\cite{SnapFu}, the shallow up/down blocks (where $l \leq 2$), as these operate on higher-resolution features, account for the majority of computational costs.

\subsubsection{Token Reduction}
Consider an input represented as a set of tokens, denoted as $\mathbf{X}=\{x_1, x_2, \dots, x_N\}$, in conjunction with a neural network $f$. The computation performed on the original input can be expressed as $f(\mathbf{X})$. The objective of token reduction is to minimize the length of $\mathbf{X}$ while preserving the accuracy of the prediction results, which can be formalized as
\begin{equation}
    \mathop{\arg\min}_{\pi} \|f(\mathbf{X}) - f(\mathbf{\pi(X)})\|,
\end{equation}
where $\pi$ represents the token reduction strategy, and $\pi(\mathbf{X})$ is the resulting reduced token set that satisfies the condition $|\pi(\mathbf{X})| \leq |\mathbf{X}|$, with $|\cdot|$ indicating the cardinality of the set. In this context, token reduction implies that $\pi(\mathbf{X}) \subset \mathbf{X}$ is a subset of the original set of tokens. For the token merging approach, $\pi(\mathbf{X}) = \{(x_m + x_n)/2\}_{m \neq n}$ is computed by averaging the two most similar tokens within $\mathbf{X}$.
\subsubsection{Feature Caching}
Following previous caching methods DeepCache~\cite{sd-deepcache}, we introduce the naive scheme for feature caching. Given a set of $\mathcal{N}$ adjacent timesteps $\{\mathcal{T}, \mathcal{T}-1, \ldots, \mathcal{T}-\mathcal{N}+1\}$ and cache depth $d$, native feature caching performs the full computation and cache outputs at the first timestep $\mathcal{T}$:
\begin{equation}
\operatorname{Cache}(l,s)=f_l^s(\mathbf{X}_{\mathcal{T}}) \quad l\geq d, s \in\{\text{up, down, mid}\} 
\end{equation}
The subsequent timesteps $t\in \{\mathcal{T}-1, \ldots, \mathcal{T}-\mathcal{N}+1$\}, reuse cached values:
\begin{equation}
f_l^s(\mathbf{X}_t)=\operatorname{Cache}^s(l) \quad l \geq d, s \in\{\text{up, down, mid}\}
\end{equation}

\subsection{Dynamic-aware Token Pruning}

\subsubsection{Base Token Selection}
\paragraph{Temporal Noise Difference Score}

For the \( t \)-th timestep, we obtain outputs \( f_0^{\text{up}}(x_{t+2}) \) and \( f_0^{\text{up}}(x_{t+1}) \) for the the two preceding adjacent timesteps, both with dimensions \( (B, C, \sqrt{N}, \sqrt{N}) \) where \( B \), \( C \), and \( N \) are the batch size, number of channels, and total tokens, respectively. We then calculate the absolute difference between these outputs and average this difference over the channels, defining it as the \textbf{DiffScore} to quantify the change across timesteps:
\begin{equation}
    \text{DiffScore} = \frac{1}{C} \sum_{c=1}^{C} \left| f_0^{\text{up}}(x_{t+2})_{c} - f_0^{\text{up}}(x_{t+1})_{c} \right|
\end{equation}
where \( f_0^{\text{up}}(x_{t+2})_{c} \) and \( f_0^{\text{up}}(x_{t+1})_{c} \) denote the values at timestep \( t+2 \) and \( t+1 \), respectively, for each channel \( c \) in the output of the \( f_0^{\text{up}} \) function. The summation averages the absolute differences over the channel dimension \( C \).

\paragraph{Patch-based Token Selection}
We then identify the base tokens in each patch of the image. Instead of directly selecting the tokens with the highest DiffScore in the whole input data, we further add the constraint that the base token should be uniformly distributed in the spatial positions on the image. Hence, for each patch in the image, we select the token with the highest DiffScore as the base token in this patch. 
Specifically, we first reshape the similarity score matrix DiffScore into a 2D grid format with dimensions $(B, \sqrt{N}, \sqrt{N})$. This transformation maps the sequence of tokens into their corresponding patches in an image.
Subsequently, we partition the grid into non-overlapping patches of size $s \times s$. Specifically, the matrix is divided into ${\sqrt{N}}/{s} \times {\sqrt{N}}/{s}$ patches, resulting in each patch encompassing $s \times s$ tokens. 
Within each patch, we select the token with the maximum similarity score in DiffScore as the base token in this patch. Hence, we obtain a sequence of base tokens which can be denoted as $\mathbf{X}_{\text{base}} \subset X$ with $|\mathbf{X}_{\text{base}}| = {|\mathbf{X}|}/{s^2}$, where its i$_{th}$ element $\mathbf{X}_{\text{base},i}$ can be formulated as 
\begin{equation}
\label{equ:simscore}
\begin{aligned}
    \mathbf{X}_{\text{base},i,j}& = \mathop{\arg\max}_{\mathbf{X}_{m,n}} \text{DiffScore}(\mathbf{X}_{m,n})
    \\ \text{s.t.}& ~~~m\in [is, (i+1)s-1],n\in [js, (j+1)s-1]
\end{aligned}
\end{equation}

\noindent\textbf{Align Base Tokens with and without CFG Guidance}
In diffusion models, the front \( B/2 \) samples are for conditional generation based on guiding conditions, while the back \( B/2 \) samples are for unconditional generation, created without specific conditions.
This alignment of base tokens for conditional and unconditional generation is grounded in experimental findings that indicate a notable enhancement in the quality of the generated images.
To align the base tokens for conditional and unconditional generation, we further define the base token tensor \( \mathbf{X}_{\text{base}, i, j} \in \mathbb{R}^{B \times \sqrt{N} \times \sqrt{N}} \) such that:

\begin{equation}
\begin{aligned}
    \mathbf{X}_{\text{base}, i, j}[k] = \mathbf{X}_{\text{base}, i, j}[k - B/2],\\ 
    \text{s.t.}~~~ k \in \{B/2, B/2 + 1, \dots, B - 1\}
\end{aligned}
\end{equation}
where the first \( B/2 \) entries are duplicated into the latter \( B/2 \) entries. Here, \( B \) is the batch size and \( N \) represents the total number of tokens.

\subsubsection{Pruned Token Selection}
Based on the base tokens, we then select the tokens to be pruned from the left tokens $\mathbf{X}-\mathbf{X}_{base}$. 
For each token inside this set, we get the highest similarity between this token and all the base tokens as the criterion for whether this token should be pruned. The tokens that are more similar to the base tokens will be pruned since recovering them from base tokens leads to lower reconstruction errors. This process can be formulated as 
\begin{equation}
\label{equ:prune}
    \begin{aligned}
    \mathbf{X}_{\text{prune}} = &\mathop{\arg \text{top}K}_{\mathbf{Xi}}  \max_{\mathbf{X}_j} \text{Cosine Similarity}({\mathbf{X}_i,\mathbf{X}_j})
    \\
    & \mathbf{X}_i \in \mathbf{X}-\mathbf{X}_{\text{base}} \text{~~~and~~~} \mathbf{X}_j \in \mathbf{X}_{\text{base}}.
    \end{aligned}
\end{equation}
where $K$ denotes the number of pruned tokens.
Please note that these cosine similarities have been previously computed in the stage of base token selection in CosSim. Hence, the cosine similarity in Eq.~\ref{equ:prune} can be directly obtained and does not require re-computation. 

\subsubsection{Pruned Token Recovery}
The pruned token sequence $\mathbf{X}-\mathbf{X}_{\text{prune}}$ is then input to the attention layers for diffusion models for computing, resulting in an output sequence $f(\mathbf{X}-\mathbf{X}_{\text{prune}})$ with the same sequence length. Then, we recover the pruned tokens by directly copying them from their most similar base tokens found in Equation~\ref{equ:prune}. 

\subsection{Timestep-aware Feature Caching}
\begin{algorithm}[t]
\caption{ Step-aware Evolutionary Searching}\label{alg:search}
\begin{algorithmic}[1]
\State {\bfseries Input: the cache depth $d$, the pruning ratio $r$, number of generation $G$, population size $P$, mutation probability $m$, latency objective weight $w_{M}$, source dataset $X$} 
\State Generate the initial population p of strategies
\For {g in range(G)}
    \For{each individual i in p}
        \State score(i) = FID(i, d, r, $X$) + $w_{M}$ $\times$ Latency(i)
        \State assign a rank to the individual i
    \EndFor
    \State selection(p, P)\,\#\textit{\,select new population}
    \State crossover(p)\,\#\textit{\,crossover among individuals}
    \State mutation({p},\,{m})\,\#\textit{\,mutate individuals according to m}
\EndFor
\State find the best individual $F$ with the highest score
\State {\bfseries Output: the best step-aware strategy $F$} 
\end{algorithmic}
\end{algorithm}

\subsubsection{Search Space Pruning}
Through preliminary experiments, we identified key factors to narrow the search space, optimizing both computational efficiency and output quality.
We found that cache depth $d$ greater than 1 incurred excessive computational costs and compromised generation quality. Therefore, we constrained $d$ to the values $\{0, 1, 12\}$ , with $d = 12$ representing complete computation without caching. 
Similarly, analysis of the pruning ratio $r$ showed that values too small yielded minimal acceleration, while larger values caused significant degradation in image quality. Based on these findings, we limited $r$ to the set $\{0.3, 0.4, 0.5, 0.6, 0.7\}$.

\subsubsection{Dynamic Caching Depth and Pruning Ratio}
Rather than performing a complete recalculation in the reverse distribution at every timestep, we leverages the function $F(t)$ to adjust the depth of the cache $d$ and the prune ratio $r$ dynamically. 
\begin{equation}
p_{F(t)}\left(x_{t-1} \mid x_t\right)=\mathcal{N}\left(x_{t-1} ; \mu_{F(t)}\left(x_t, t\right), \Sigma_{F(t)}\left(x_t, t\right)\right),
\end{equation}
To identify the optimal strategy $F$, we employ a multi-objective evolutionary search focused on minimizing model size while maximizing performance across various computational constraints. We leverage NSGA-II~\cite{nsga2}, an efficient multi-objective optimization algorithm renowned for its ability to balance convergence and diversity. This algorithm ranks solutions using non-dominated sorting, forming fronts that prioritize the best solutions and preserve diversity through crowding distance.

Our search process is initialized with a mix of uniform non-step-aware strategies and random strategies in the first generation. To further enhance diversity, we utilize single-point crossover at random positions and apply random choice mutations, which adjust the model size of a selected step to another available option. The specifics of this search algorithm are detailed in \cref{alg:search}. Once we have identified the optimal step-aware strategy $F$, we can leverage it to accelerate the sampling process. During each sampling iteration, our strategy $F$ is utilized to determine the most appropriate pruning ratio $r$ and cache depth $d$.
 \section{Experimental Results}
\begin{figure}[t]
\centering
\includegraphics[width=\linewidth]{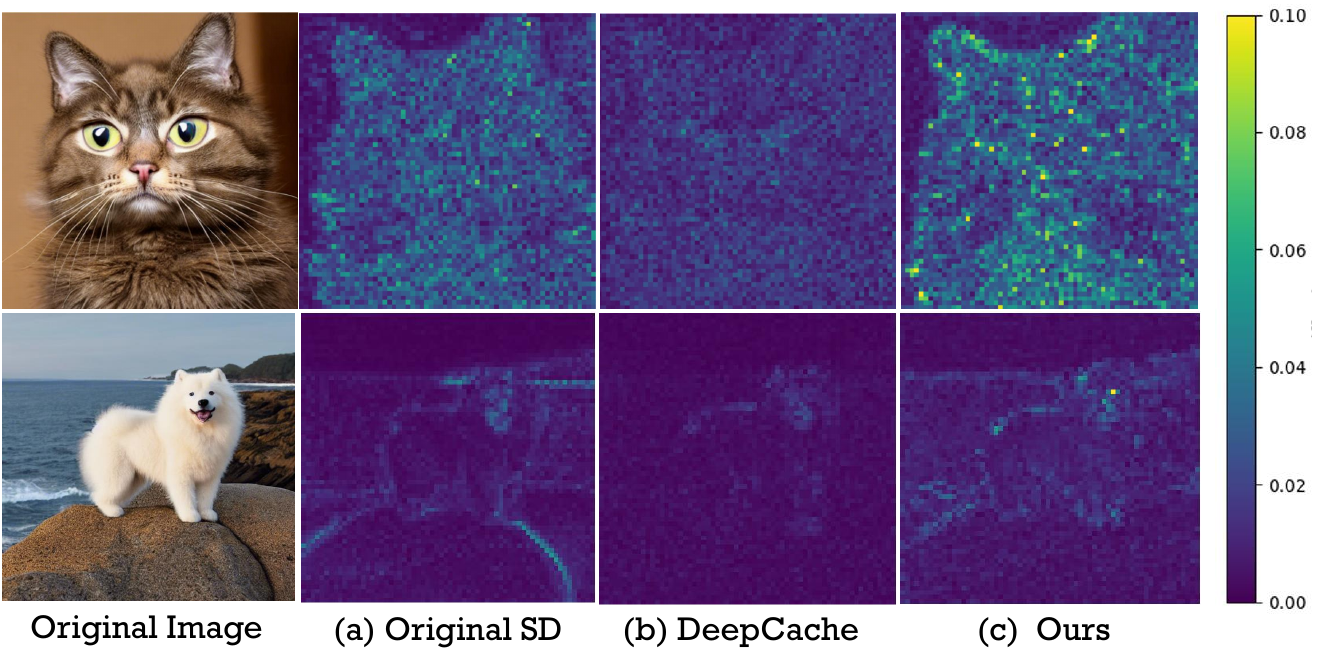} 

\caption{
Visualization of feature heatmaps of the difference value between adjecent timesteps: original SD (without acceleration), cache-only, and our method. \textbf{(a):} The original SD. \textbf{(b)}: DeepCache\cite{DeepCache} reduces feature dynamics across timesteps, resulting in a loss of semantic information. \textbf{(c)}: Our method effectively restores rich semantic information in features, preserving generation quality while boosting efficiency.
}
\vspace{-10pt}
\label{fig:heatmap}
\end{figure}
\label{sec:experimental_results}
\begin{table*}[t]
\caption{Comparison on Stable Diffusion v1.5 and v2 using ImageNet and COCO30k datasets. We evaluate each method based on FID\cite{fid}, average latency for generating each image (in seconds), and speedup ratio. The table includes multiple configurations (e.g., a1, b1, etc.) representing different model parameter settings with varying levels of acceleration. $\uparrow$ and $\downarrow$: lower and higher values are better.}
\vspace{-5pt}
\centering
\resizebox{\linewidth}{!}{
\begin{tabular}{ccccc|cccc}
\toprule
\multirow{2}{*}{\textbf{Dataset}}   & \multicolumn{4}{c}{\textbf{Stable Diffusiion v1.5}}                                                                                                     & \multicolumn{4}{c}{\textbf{Stable Diffusion v2}}                                                                                                  \\ \cmidrule{2-5} \cmidrule{6-9}
                                    & \textbf{Method}                 & \textbf{FID$\downarrow$}   & \begin{tabular}[c]{@{}c@{}}\textbf{Latency$\downarrow$} \\ \textbf{(Second)}\end{tabular} & \begin{tabular}[c]{@{}c@{}}\textbf{Speedup $\uparrow$} \\ \textbf{Ratio}\end{tabular}  & \textbf{Method}   & \textbf{FID$\downarrow$}   & \begin{tabular}[c]{@{}c@{}}\textbf{Latency$\downarrow$}\\ \textbf{(Second)}\end{tabular} & \begin{tabular}[c]{@{}c@{}}\textbf{Speedup} \\ \textbf{Ratio} $\uparrow$\end{tabular} \\ \hline
\multirow{14}{*}{\textbf{ImageNet}} & Original SD & 27.64 & 2.61                                                        & 1.00& Original SD       & 29.8  & 2.61                                                       & 1.00\\ \cline{2-9} 
                                    & DDIM (10 steps)        & 27.80& 0.60& 4.34                                                     & DPM (10 Steps)    & 29.48 & 0.75                                                       & 3.48                                                     \\ \cline{2-9} 
                                    & ToMeSD-a1                 & 27.49 & 1.39                                                        & 1.88                                                     & ToMeSD-a2           & 29.2  & 1.66                                                       & 1.57                                                     \\
                                    & ToMeSD-b1                 & 27.99 & 1.27                                                        & 2.05                                                     & ToMeSD-b2            & 29.41 & 1.57                                                       & 1.66                                                     \\ \cline{2-9} 
                                    & DeepCache-a1              & 29.36 & 0.59                                                        & 4.40& DeepCache-a2         & 30.75 & 1.05                                                       & 2.49                                                     \\
                                    & DeepCache-b1              & 35.37 & 0.37                                                        & 7.10& DeepCache-b2         & 35.96 & 1.00& 2.61                                                     \\ \cline{2-9} 
                                    & DeepCache\&ToMeSD-a1      & 27.58 & 0.49                                                        & 5.32                                                     & DeepCache\&ToMeSD-a2 & 30.62 &                                                            0.47&                                                          5.55\\
                                    & DeepCache\&ToMeSD-b1      & 28.87 & 0.40& 6.50& DeepCache\&ToMeSD-b2 & 29.08 &                                                            0.44&                                                          5.93\\
                                    & DeepCache\&ToMeSD-c1      & 30.92 & 0.36                                                        & 7.27                                                     & DeepCache\&ToMeSD-c2 & 29.19 &                                                            0.41&                                                          6.37\\ \cline{2-9} 
                                    & Ours-a1                 & 25.20  & 0.65                                                        & 4.03                                                     & Ours-a2              & 26.44 & 0.63                                                       & 4.14                                                     \\
                                    & Ours-b1                   & 25.40  & 0.45                                                        & 5.75                                                     & Ours-b2              & 27.35 & 0.59                                                       & 4.41                                                     \\
                                    & Ours-c1                   & 25.70  & 0.37                                                        & 7.00& Ours-c2              & 27.58 & 0.44                                                       & 5.97                                                     \\
                                    & Ours-d1                   & 26.89 & 0.30& 8.61                                                     & Ours-d2              & 27.83 & 0.41                                                       & 6.34                                                     \\
                                    & Ours-e1                  & 27.31 & 0.29                                                  & 9.01& Ours-e2 & 28.20  &0.36            & 7.25 \\ \midrule
\multirow{9}{*}{\textbf{COCO30k}}   & Original SD            & 12.15 & 2.61                                                        & 1.00& Original SD       & 13.68 & 2.46                                                       & 1.00\\ \cline{2-9} 
                                    & DDIM (10 Steps)        & 12.60  & 0.60& 4.34                                                     & DPM (10 Steps)    & 14.35 & 0.75                                                       & 3.48                                                     \\ \cline{2-9} 
                                    & DeepCache-d1              & 16.74 & 0.64                                                        & 4.40& DeepCache-d2         & 21.56 &                                                            0.76&                                                          3.24\\
                                    & DeepCache-e1              & 26.57 & 0.37                                                        & 7.10& DeepCache-e2         & 96.79 & 0.77& 3.19\\ \cline{2-9} 
                                    & DeepCache\&ToMeSD-d1      & 11.53 & 0.49                                                        & 5.32                                                     & DeepCache\&ToMeSD-d2 & 18.82 & 0.47& 5.23\\
                                    & DeepCache\&ToMeSD-e1      & 12.84 & 0.40& 6.50& DeepCache\&ToMeSD-e2 & 17.54& 0.43& 5.72\\ \cline{2-9} 
                                    & Ours-f1                  & 9.35  & 0.57                                                        & 4.55                                                     & Ours-a2              & 10.12 & 0.63                                                       & 4.14                                                     \\
                                    & Ours-g1                   & 9.41  & 0.49                                                        & 5.33                                                     & Ours-f2              & 10.77 & 0.51                                                       & 5.08                                                     \\
                                    &  Ours-c1                  & 9.98  &0.37                                                       &7.00 & Ours-e2              & 13.88 & 0.36                                                       & 7.25                                                    \\ \bottomrule
\end{tabular}
}
\vspace{-5pt}
\label{tab:all_compare}
\end{table*}

\subsection{Experimental Setup}
\paragraph{Evaluation}
Our experiments are conducted with SD v1.5 and SD v2 by generating 512$\times$512 images using 50 PLMS~\cite{plms} steps with a cfg scale~\cite{dm-3} of 7.5 and 9.0, respectively.
We generate 2,000 images of ImageNet-1k~\cite{imagenet} (2 per class) and 30,000 images of COCO30k~\cite{coco} classes (1 per caption) for evaluation. 
For SDXL, the models generate 1024$\times$1024 pixel images based on captions from 5k images in the MS COCO validation set\cite{coco}. The default configuration includes a CFG scale of 7.0, 50 sampling steps, and the EulerEDMSampler.
FID~\cite{fid}, Inception Score~\cite{isscore} and CLIP Score~\cite{clipscore} are utilized as the metrics for generation quality.
The average latency for generating an image and speedup are measured on a single 4090 GPU.

\noindent\textbf{Search Setting}
We employed NSGA-II~\cite{nsga2} for our search algorithm, implementing it using the open-source libraries pymoo~\cite{pymoo} and PyTorch~\cite{pytorch}. 
We only perform searching in Stable Diffusion v1.5 on 500 ImageNet images and transfer the obtain strategy in all other datasets and models.

\subsection{Qualitative Analysis}
As shown in Fig.\ref{fig:overall_visualization}, we evaluate DaTo by using manually crafted challenging prompts.
We benchmarked our DaTo method against with other training-free acceleration methods, including token merge method (ToMeSD~\cite{sd-tome}), feature cache method (DeepCache~\cite{DeepCache}) and the naive combination of these two approaches (DeepCache \& ToMeSD). The visual comparison results highlight the advantages of DaTo in the following aspects:

\noindent\textbf{Content Fidelity and Detail Preservation}: We evaluate the model on SDXL text-to-image generation by generating $1024\times 1024$ size images.
As illustrated in \cref{fig:overall_visualization} (top left quadrant), DaTo’s output closely resembles the original image content (\textit{\eg~the contours of the main subject}) while achieving a 2$\times$ speed-up over SDXL.
In high-detail scenes, DaTo retained intricate textures better than other method, as shown in \cref{fig:overall_visualization} (top right quadrant).

\noindent\textbf{High-Quality Style Adaptation}: For SD1.5 image-to-image tasks (See \cref{fig:overall_visualization}, bottom left quadrant), DaTo excelled in style adaptation, balancing content preservation with style accuracy without harming essential details. 

\noindent\textbf{Prompt Alignment}: DaTo also excelled at aligning with complex text prompts in SD1.5 txt2img tasks. As shown in \cref{fig:overall_visualization} (bottom right quadrant), DaTo achieved 9$\times$ speedup with better prompt alignment accuracy than other methods. For example, only our methods generate all the elements in the given prompts correctly, such as \textit{farmhouse, howling wolf,} and \textit{full moon}.

\subsection{Quantitative Evaluations}

\subsubsection{Evaluation on the Stable Diffusion Model}
\cref{tab:all_compare} presents a comprehensive comparison of DaTo with other training-free acceleration techniques, including DDIM~\cite{ddim}, ToMeSD~\cite{sd-tome}, DeepCache~\cite{DeepCache} and naive combination of ToMeSD and DeepCache (ToMeSD \& DeepCache).
We conduct experiments on Stable Diffusion v1.5 and v2, evaluated on the ImageNet~\cite{imagenet} and COCO30k~\cite{coco} datasets. We achieve varying levels of speedup through different settings of cache depth or prune ratio, represented as -a1, -a2, etc. These settings are determined by searching for the optimal configuration in SDv1.5, which is then directly applied to evaluate generalization performance on other models and datasets. More details about the configurations are provided in the appendix. DaTo consistently achieves lower FID scores and higher speedup ratios across all configurations. Specifically, on ImageNet with SD v1.5, our configuration e1 achieves a 9.01 speedup while maintaining a lossless FID of 27.31. For SD v2, our configuration e2 delivers a speedup of 7.25, accompanied by a FID of 28.20.
In COCO30k, our configuration h1 on SD v1.5 attains a 7$\times$ speedup with an FID reduction of 2.17 than the original SD, while our configuration h2 on SD v2 achieves a 7.25$\times$ speedup, with an FID of 13.88.
These results highlight the effectiveness of DaTo in maintaining high image quality while substantially reducing computational cost, outperforming other acceleration strategies.
 
\subsubsection{Evaluation on the Stable Difussion XL model}
\begin{table}[t]
\caption{MS COCO zero-shot evaluation on SDXL.}
\vspace{-5pt}
\centering
\resizebox{\linewidth}{!}{
\begin{tabular}{@{}cccc@{}}
\toprule
\textbf{Method}          & \textbf{FID$\downarrow$}   & \textbf{Latency (Second) $\downarrow$} & \textbf{Speedup Ratio $\uparrow$}\\ \midrule
Original SDXL   & 24.25 & 12.47   & 1.00\\
DeepCache       & 26.39 & 7.13    & 1.75    \\
 DeepCache\&TomeSD & 31.21 & 6.99    & 1.78    \\
\rowcolor[HTML]{ECF4FF} Ours    & 23.10& 5.37    & 2.32    \\ \bottomrule
\end{tabular}
}
\label{tab:sdxl}
\vspace{-5pt}
\end{table}

In \cref{tab:sdxl}, we compare the performance of our method with existing methods on the SDXL model using the MS COCO validation dataset~\cite{coco} to generate images with the size of $1024\times 1024$. DeepCache achieves a latency reduction with a speedup ratio of 1.75, though at the cost of a slightly higher FID of 26.39. DeepCache \& Tome further reduces latency, achieving a speedup of 1.78, but results in an FID increase to 31.21. In contrast, our proposed method achieves the best balance, with the lowest FID of 23.10 and the highest speedup ratio of 2.32, demonstrating both the superior quality and efficiency of our method in more advanced diffusion models and high-resolution images.

\subsection{Ablation Study}

\begin{table}[t]
\centering
\caption{Impact of the alignment of base tokens with and w/o cfg guidance under different cache settings. We report the results on FID scores (lower FID is better).}
\label{tab:aligncfg}
\vspace{-5pt}
\resizebox{\linewidth}{!}{
\begin{tabular}{ccccc}
\hline
\multirow{2}{*}{\textbf{\begin{tabular}[c]{@{}c@{}}Pruning \\ Ratio\end{tabular}}} & \multicolumn{2}{c}{\textbf{Cache setting 1}} & \multicolumn{2}{c}{\textbf{Cache setting 2}} \\
                                                                                  & \textbf{w/ alignment}    & \textbf{w/o alignment}    & \textbf{w/ alignment}    & \textbf{w/o alignment}   \\ \hline
0.4                                                                             & \cellcolor[HTML]{ECF4FF}{\color[HTML]{000000} 25.44}                  & 25.64                 &  \cellcolor[HTML]{ECF4FF}{\color[HTML]{000000}25.77}                  & 27.84                \\
0.5                                                                               &  \cellcolor[HTML]{ECF4FF}{\color[HTML]{000000}25.86}                  & 27.15                 &  \cellcolor[HTML]{ECF4FF}{\color[HTML]{000000}25.82}                  & 31.62                \\
0.6                                                                               &  \cellcolor[HTML]{ECF4FF}{\color[HTML]{000000}26.41}                  & 30.89                 &  \cellcolor[HTML]{ECF4FF}{\color[HTML]{000000}25.85}                  & 38.92                \\
0.7                                                                               &  \cellcolor[HTML]{ECF4FF}{\color[HTML]{000000}28.17}                  & 41.26                 & \cellcolor[HTML]{ECF4FF}{\color[HTML]{000000}27.01}                  &  57.45                \\ \hline
\end{tabular}
}
\vspace{-5pt}
\end{table}
\begin{table}[t]
\centering
\caption{Impact of Temporal Noise Difference Score on FID across pruning ratios and cache settings (lower FID is better).}
\vspace{-5pt}
\label{tab:diff}
\resizebox{\linewidth}{!}{
\begin{tabular}{@{}ccccc@{}}
\hline
\toprule
\multirow{2}{*}{\textbf{\begin{tabular}[c]{@{}c@{}}Pruning \\ Ratio\end{tabular}}} & \multicolumn{2}{c}{\textbf{Cache setting 1}}   & \multicolumn{2}{c}{\textbf{Cache setting 2}}   \\
                                                                                  & \textbf{w/ DiffScore} & \textbf{w/o DiffScore} & \textbf{w/ DiffScore} & \textbf{w/o DiffScore} \\ \midrule
0.3                                                                               & \cellcolor[HTML]{ECF4FF}{\color[HTML]{000000}25.78}                   &  25.94                   & \cellcolor[HTML]{ECF4FF}{\color[HTML]{000000} 26.44}                   & 26.74                  \\
0.4                                                                               & \cellcolor[HTML]{ECF4FF}{\color[HTML]{000000} 26.01}                   & 26.07                  & \cellcolor[HTML]{ECF4FF}{\color[HTML]{000000} 26.68}                   & 26.94                  \\
0.5                                                                               & \cellcolor[HTML]{ECF4FF}{\color[HTML]{000000} 26.19}                   & 26.41                  & \cellcolor[HTML]{ECF4FF}{\color[HTML]{000000}26.93}                   &  27.05                  \\
0.6                                                                               & \cellcolor[HTML]{ECF4FF}{\color[HTML]{000000} 26.86}                   & 27.23                  & \cellcolor[HTML]{ECF4FF}{\color[HTML]{000000} 27.76}                   & 28.11                  \\
0.7                                                                               & \cellcolor[HTML]{ECF4FF}{\color[HTML]{000000} 28.74}                   & 28.81                  & \cellcolor[HTML]{ECF4FF}{\color[HTML]{000000}29.91 }                  &  29.93                  \\ \bottomrule
\end{tabular}
}
\end{table}
\begin{table}[t]
\centering
\caption{Comparison of CLIP Score~\cite{clipscore}, Inception Score~\cite{isscore}, Latency, and Speedup Ratio for different methods on SD1.5 image-to-image generation. $\uparrow$ and $\downarrow$: lower and higher values are better.}
\vspace{-5pt}
\resizebox{\linewidth}{!}{
\begin{tabular}{@{}ccccc@{}}
\toprule
\textbf{Method}      & \textbf{CLIP Score$\uparrow$} & \textbf{Inception Score $\uparrow$} & \begin{tabular}[c]{@{}c@{}}\textbf{Latency}\\ \textbf{(Second)$\downarrow$}\end{tabular} & \begin{tabular}[c]{@{}c@{}}\textbf{Speedup}\\ \textbf{Ratio$\uparrow$}\end{tabular} \\ \midrule
Original SD & 17.45      & 42.91           & 11.09                                                      & 1.0                                                     \\
DeepCache   & 17.45      & 42.88           & 4.22                                                       & 2.63                                                    \\

\rowcolor[HTML]{ECF4FF}Ours        & 17.48      & 42.98           & 2.80                                                       & 3.96                                                    \\ \bottomrule
\end{tabular}
}
\label{tab:i2i}
\vspace{-5pt}
\end{table}

\subsubsection{Effectiveness of DiffScore}
As shown in \cref{tab:diff}, incorporating DiffScore consistently improves FID with two kinds of cache strategy, as demonstrated in experiments on SDv1.5 with ImageNet1k. 
Additionally, ~\cref{fig:heatmap} shows that tokens with higher temporal differences between adjacent timesteps are identified as the most crucial parts of the generation process. By focusing on these high-variance tokens, DaTo enhances the fidelity of key details and improves overall generation results.

\subsubsection{The Alignment with and w/o CFG Guidance}
To evaluate the impact of alignment of base tokens between conditional and unconditional generation, the experiment was conducted on 2,000 images generated from the ImageNet dataset, with the FID score used to quantify image quality. As shown in ~\cref{tab:aligncfg}, the alignment configuration yields progressively greater benefits as the pruning ratio increases under two kinds of cache strategy, significantly enhancing the quality of generated images.

\subsection{Performance of Our Method in Other Metrics}
\cref{tab:i2i} shows a performance comparison between Original SD, DeepCache, and our proposed method for SD1.5 image-to-image generation. Each method generates 1,000 1024$\times$1024 images from ImageNet-1k (one per class), where original images are generated by the SD1.5 model at 512$\times$512 resolution and then resized to 1024$\times$1024.
We report CLIP Score~\cite{clipscore} with CLIP-ViT-g/14 model and Inception Scores~\cite{isscore} for image quality, Latency (in seconds) and Speedup Ratio for processing speed. Our method achieves the highest speedup ratio of 3.96, significantly reducing latency while maintaining or slightly improving image quality compared to the baseline.

 \section{Conclusion}
\label{sec:conclusion}
In this work, we introduce DaTo (Dynamics-Aware Token Pruning), a hardware-friendly and training-free token pruning approach that mitigates the quality drop typically caused by feature caching. DaTo prunes tokens whose dynamics across timesteps are diminished by feature caching and recovers them using tokens with larger dynamics. Furthermore, we propose an evolutionary method to search for the optimal feature caching and token pruning strategy, fully unlocking the potential of DaTo. 
DaTo achieves a 9$\times$ speedup on ImageNet and 7$\times$ speed up on COCO-30k, with lossless image quality, demonstrating the it's significant effectiveness in improving both computational efficiency and image quality.
\clearpage
\setcounter{page}{1}
\maketitlesupplementary

\section{Detailed Experimental Settings}
To address the optimization problem, we employ the Non-dominated Sorting Genetic Algorithm II (NSGA-II)~\cite{nsga2}, a widely-used approach for solving multi-objective optimization problems. In our setup, the population size is set to 20, meaning that each generation consists of 20 candidate solutions. The initial population is generated using IntegerRandomSampling, ensuring that all solutions are represented as integers. For crossover, we use Simulated Binary Crossover (SBX) with a probability of 0.7 and a distribution index of 7, which controls the diversity of offspring solutions. Mutation is performed using Polynomial Mutation (PM), with a mutation probability of 0.4 and a distribution index of 15 to regulate the magnitude of mutations. Additionally, duplicate solutions are eliminated in each generation to maintain population diversity. The optimization process terminates after 100 generations. We execute the optimization using the minimize function, which integrates the problem definition, algorithm configuration, and termination conditions. 
Additionally, we adopt two objectives: FID~\cite{fid} for measuring generation quality and Latency for computational efficiency, balanced by a weight factor $w_m$. A higher $w_m$ (e.g., 0.16) places more emphasis on minimizing latency, while a lower $w_m$ (e.g., 0.025) prioritizes quality.

To further illustrate our approach, we take the Ours-b1 setting as an example. Our search strategy determines the optimal configuration of pruning ratios and cache IDs for each step in the optimization process. Specifically, for every step, we explore the search space to identify the best combination of pruning ratio and cache ID. The detailed numerical results corresponding to the Ours-b1 setting are shown in \cref{tab:app_search}.

\begin{table*}[t]
\centering
\caption{Specific search strategy corresponding to the “Ours-b1” setting.}
\resizebox{\linewidth}{!}{
\begin{tabular}{@{}lccccccccccccccccccccccccc@{}}
\toprule
\textbf{Timestep}    & 49  & 48  & 47  & 46  & 45  & 44  & 43  & 42  & 41  & 40  & 39  & 38  & 37  & 36  & 35  & 34  & 33  & 32  & 31  & 30  & 29  & 28  & 27  & 26  & 25  \\ \midrule
\textbf{Cache ID} & 1   & 1   & 1   & 1   & 1   & 1   & 1   & 1   & 12  & 1   & 1   & 1   & 1   & 1   & 1   & 1   & 1   & 0   & 0   & 12  & 1   & 1   & 1   & 1   & 1   \\
\textbf{Pruning Ratio} & 0.7 & 0.5 & 0.4 & 0.4 & 0.3 & 0.5 & 0.6 & 0.5 & 0.6 & 0.5 & 0.5 & 0.3 & 0.6 & 0.5 & 0.5 & 0.4 & 0.7 & 0.3 & 0.6 & 0.5 & 0.5 & 0.3 & 0.4 & 0.3 & 0.4 \\ \midrule
\textbf{Timestep}  & 24  & 23  & 22  & 21  & 20  & 19  & 18  & 17  & 16  & 15  & 14  & 13  & 12  & 11  & 10  & 9   & 8   & 7   & 6   & 5   & 4   & 3   & 2   & 1   & 0   \\
\textbf{Cache ID}      & 0   & 1   & 12  & 0   & 1   & 1   & 12  & 0   & 0   & 1   & 1   & 1   & 1   & 1   & 1   & 1   & 1   & 1   & 12  & 1   & 1   & 1   & 1   & 1   & 0   \\
\textbf{Pruning Ratio} & 0.3 & 0.6 & 0.5 & 0.6 & 0.3 & 0.6 & 0.6 & 0.3 & 0.3 & 0.5 & 0.6 & 0.4 & 0.4 & 0.3 & 0.4 & 0.4 & 0.5 & 0.3 & 0.5 & 0.6 & 0.7 & 0.6 & 0.5 & 0.6 & 0.5 \\ \bottomrule
\end{tabular}
}
\label{tab:app_search}
\end{table*}

\section{Additional Qualitative Results}
\begin{figure}[t]
\centering
\includegraphics[width=\linewidth]{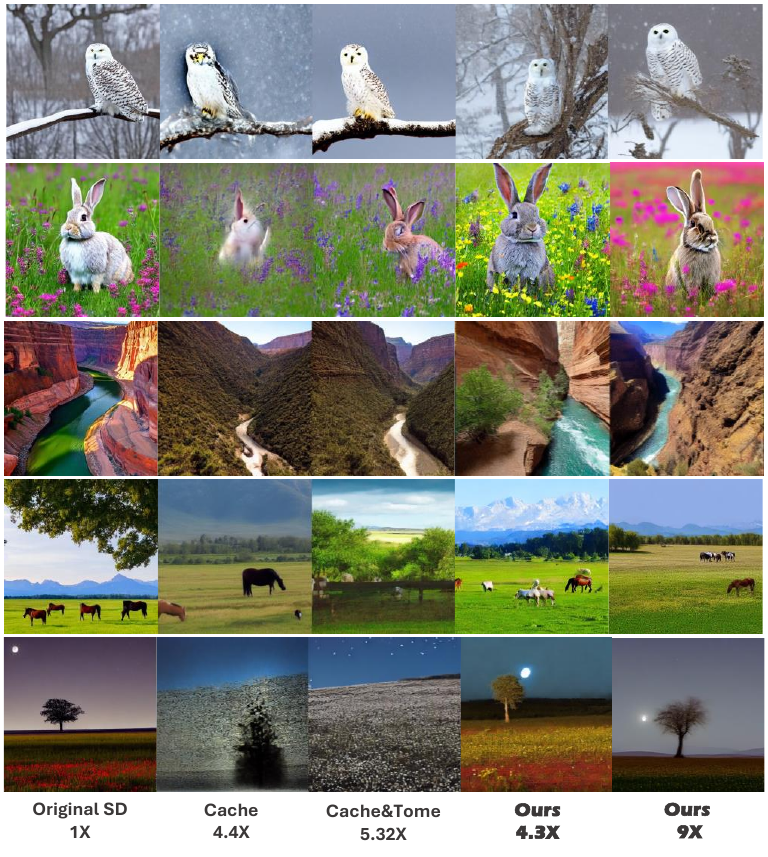}
\vspace{-0.6cm}
\caption{
Visual comparison on SD1.5 across methods, including caching only (DeepCache~\cite{DeepCache}), token pruning only (ToMeSD~\cite{sd-tome}), and a naive combination of both (DeepCache \& ToMeSD).
}
\vspace{-0.3cm}
\label{fig:app_sd}
\end{figure}

\begin{figure*}[t]
\centering
\small
\includegraphics[width=1\textwidth]{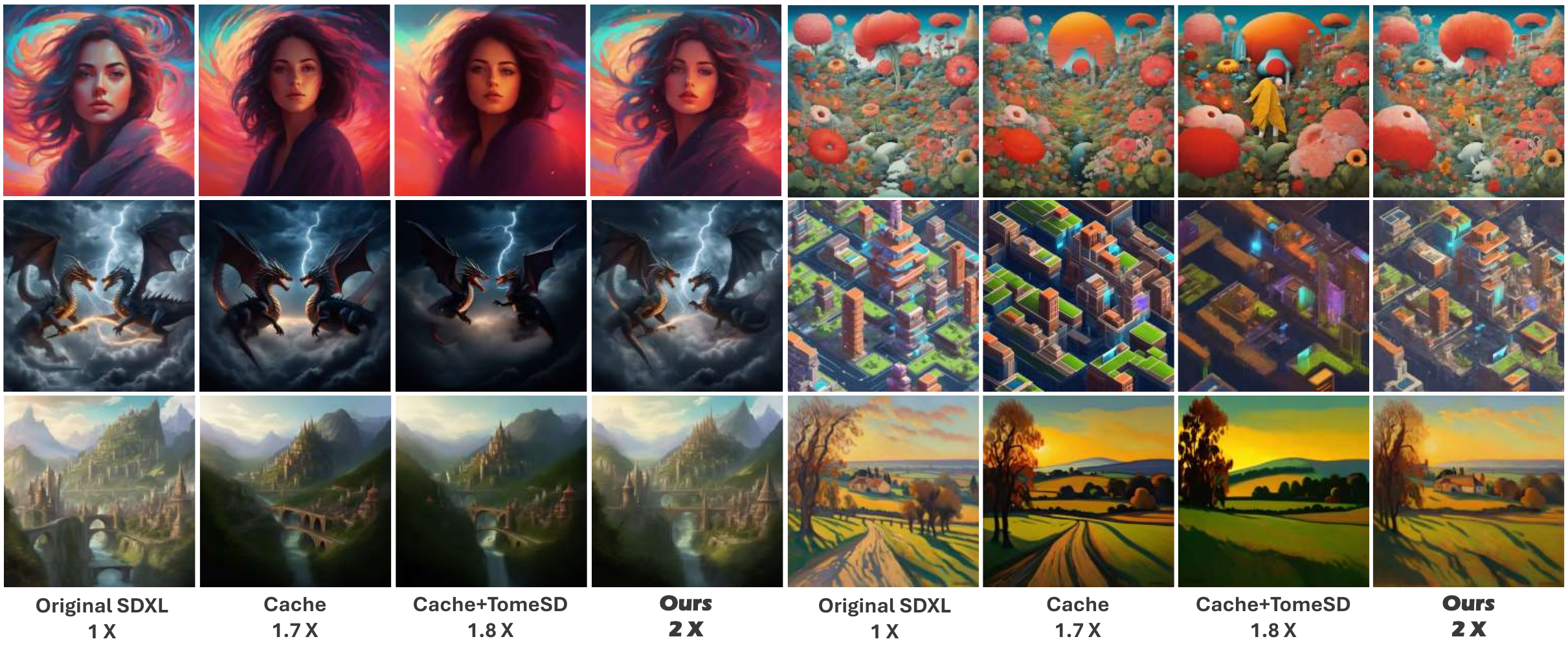}
\vspace{-20pt}
\caption{Visual comparison on SDXL across methods, including caching only (DeepCache~\cite{sd-deepcache}), token pruning only (ToMeSD\cite{sd-tome}), and a naive combination of both (DeepCache \& ToMeSD).}

\label{fig:app_sdxl}
\end{figure*}

We provide more visualization results to demonstrate the effectiveness of our proposed method, DaTo, when applied to both SD1.5 and SDXL. As shown in \cref{fig:app_sd} and \cref{fig:app_sdxl}, our method consistently achieves superior results compared to other approaches.
\section{Understanding the FID Reduction under Accelerated Conditions}
Intuitively, it seems reasonable to assume that accelerated conditions, which often result in information loss, would negatively impact performance.
However, our experimental results show that DaTo, along with some previous acceleration methods such as ToMeSD\cite{sd-tome}, achieves lower FID values compared to the original unaccelerated model under certain experimental settings.
This phenomenon can be explained through the principles of sparse coding theory. 
\paragraph{Sparse Coding Perspective.}  
Sparse coding seeks to represent a signal as a sparse combination of dictionary bases, formulated as:  

\begin{equation}
   \min _{D, X} \|Y - DX\|_F^2 \quad \text{s.t.} \quad \forall i, \|\boldsymbol{x}_i\|_0 \leq T_0    
\end{equation}  

Here, \( Y \) represents the original signal, \( D \) is the dictionary consisting of base signals (or atoms), and \( \boldsymbol{x}_i \) is the sparse representation of each token in \( X \). The optimization objective is to minimize the reconstruction error, measured as the Frobenius norm \( \|Y - DX\|_F^2 \), while ensuring that each representation \( \boldsymbol{x}_i \) has a sparsity level constrained by \( T_0 \). This approach effectively balances reconstruction accuracy and representational simplicity, promoting a compact and efficient representation of the original signal by focusing on its most salient features.
\paragraph{Token Pruning and Feature Caching is a Special Case of Sparse Coding.}
We suggest that token pruning and feature caching can be considered as a special case of sparse coding. In this context, \(Y\) represents the original input features or tokens before token pruning or feature caching, \(D\) denotes the combined set of base tokens and cached feature representations used as a dictionary to approximate the input, and \(X\) represents the sparse coefficients or mappings that reconstruct the input features \(Y\) using the elements in \(D\), including both the base tokens and the cached features.
DaTo achieves an effective balance between temporal and token-wise information reuse, maintaining sparsity in $X$ while enhancing the temporal dynamics captured in $D$. This integration of pruning and caching not only minimizes reconstruction loss but also extends feature dynamics across timesteps, leading to a more dynamic and efficient representation aligned with sparse coding principles.
\paragraph{Token Pruning and Feature Caching (Spase Coding) Can Improve Model Performance.}
Sparse coding improves model performance by focusing on essential data components and minimizing noise, as shown in Olshausen and Field’s work~\cite{hoyer2002natural}, which demonstrated enhanced fidelity through sparse representations. 
Similar effects have been observed with moderate sparsity in knowledge distillation~\cite{pointdistiller} and parameter pruning~\cite{deepcompression}.

\paragraph{Token Pruning and Feature Caching Harm Performance under Extreme Pruning and Infrequent Updates.}

Token pruning and feature caching, while effective for improving efficiency, can degrade model performance when the pruning ratio becomes excessively high or the cache update frequency is very low. In such scenarios, the model may lose access to critical dynamic features or rely too heavily on outdated cached representations, leading to incomplete or inaccurate approximations of the input features. This highlights the importance of balancing pruning and caching strategies to maintain a sufficient level of feature diversity and temporal relevance for optimal model performance.

At very high pruning ratios or or under low cache update frequencies. (e.g. DeepCache\&ToMeSD-c1), sparse coding models have too few active elements in \( D \), making high-quality reconstruction from \( DX \) to \( Y \) difficult. This excessive pruning degrades image quality, resulting in higher FID scores, showing that overly aggressive pruning or caching can harm performance by discarding essential information.

{
    \small
    \bibliographystyle{ieeenat_fullname}
    \bibliography{main}

\begin{thebibliography}{46}
\providecommand{\natexlab}[1]{#1}
\providecommand{\url}[1]{\texttt{#1}}
\expandafter\ifx\csname urlstyle\endcsname\relax
  \providecommand{\doi}[1]{doi: #1}\else
  \providecommand{\doi}{doi: \begingroup \urlstyle{rm}\Url}\fi

\bibitem[Blank and Deb(2020)]{pymoo}
Julian Blank and Kalyanmoy Deb.
\newblock Pymoo: Multi-objective optimization in python.
\newblock \emph{Ieee access}, 8:\penalty0 89497--89509, 2020.

\bibitem[Bolya and Hoffman(2023)]{sd-tome}
Daniel Bolya and Judy Hoffman.
\newblock Token merging for fast stable diffusion.
\newblock In \emph{Proceedings of the IEEE/CVF conference on computer vision and pattern recognition}, pages 4599--4603, 2023.

\bibitem[Bolya et~al.(2022)Bolya, Fu, Dai, Zhang, Feichtenhofer, and Hoffman]{vit-tome}
Daniel Bolya, Cheng-Yang Fu, Xiaoliang Dai, Peizhao Zhang, Christoph Feichtenhofer, and Judy Hoffman.
\newblock Token merging: Your vit but faster.
\newblock \emph{arXiv preprint arXiv:2210.09461}, 2022.

\bibitem[Chen et~al.(2024)Chen, Shen, Ye, Cao, Tu, Bouganis, Zhao, and Chen]{DeltaDiT}
Pengtao Chen, Mingzhu Shen, Peng Ye, Jianjian Cao, Chongjun Tu, Christos-Savvas Bouganis, Yiren Zhao, and Tao Chen.
\newblock Delta-dit: A training-free acceleration method tailored for diffusion transformers.
\newblock \emph{arXiv preprint arXiv:2406.01125}, 2024.

\bibitem[Deb et~al.(2002)Deb, Pratap, Agarwal, and Meyarivan]{nsga2}
Kalyanmoy Deb, Amrit Pratap, Sameer Agarwal, and TAMT Meyarivan.
\newblock A fast and elitist multiobjective genetic algorithm: Nsga-ii.
\newblock \emph{IEEE transactions on evolutionary computation}, 6\penalty0 (2):\penalty0 182--197, 2002.

\bibitem[Deng et~al.(2009)Deng, Dong, Socher, Li, Li, and Fei-Fei]{imagenet}
Jia Deng, Wei Dong, Richard Socher, Li-Jia Li, Kai Li, and Li Fei-Fei.
\newblock Imagenet: A large-scale hierarchical image database.
\newblock In \emph{2009 IEEE conference on computer vision and pattern recognition}, pages 248--255. Ieee, 2009.

\bibitem[Dhariwal and Nichol(2021)]{dm-3}
Prafulla Dhariwal and Alexander Nichol.
\newblock Diffusion models beat gans on image synthesis.
\newblock \emph{Advances in neural information processing systems}, 34:\penalty0 8780--8794, 2021.

\bibitem[Han et~al.(2015)Han, Mao, and Dally]{deepcompression}
Song Han, Huizi Mao, and William~J Dally.
\newblock Deep compression: Compressing deep neural networks with pruning, trained quantization and huffman coding.
\newblock \emph{arXiv preprint arXiv:1510.00149}, 2015.

\bibitem[Hessel et~al.(2021)Hessel, Holtzman, Forbes, Bras, and Choi]{clipscore}
Jack Hessel, Ari Holtzman, Maxwell Forbes, Ronan~Le Bras, and Yejin Choi.
\newblock Clipscore: A reference-free evaluation metric for image captioning.
\newblock \emph{arXiv preprint arXiv:2104.08718}, 2021.

\bibitem[Heusel et~al.(2017)Heusel, Ramsauer, Unterthiner, Nessler, and Hochreiter]{fid}
Martin Heusel, Hubert Ramsauer, Thomas Unterthiner, Bernhard Nessler, and Sepp Hochreiter.
\newblock Gans trained by a two time-scale update rule converge to a local nash equilibrium.
\newblock \emph{Advances in neural information processing systems}, 30, 2017.

\bibitem[Ho et~al.(2020)Ho, Jain, and Abbeel]{dm-ddpm}
Jonathan Ho, Ajay Jain, and Pieter Abbeel.
\newblock Denoising diffusion probabilistic models.
\newblock \emph{Advances in neural information processing systems}, 33:\penalty0 6840--6851, 2020.

\bibitem[Ho et~al.(2022)Ho, Chan, Saharia, Whang, Gao, Gritsenko, Kingma, Poole, Norouzi, Fleet, et~al.]{video-1}
Jonathan Ho, William Chan, Chitwan Saharia, Jay Whang, Ruiqi Gao, Alexey Gritsenko, Diederik~P Kingma, Ben Poole, Mohammad Norouzi, David~J Fleet, et~al.
\newblock Imagen video: High definition video generation with diffusion models.
\newblock \emph{arXiv preprint arXiv:2210.02303}, 2022.

\bibitem[Hoyer(2002)]{hoyer2002natural}
Patrik~O Hoyer.
\newblock Natural image statistics and efficient coding.
\newblock In \emph{IEEE Workshop on Neural Networks for Signal Processing, 2002}, pages 557--565, 2002.

\bibitem[Kim et~al.(2023{\natexlab{a}})Kim, Song, Castells, and Choi]{OAC}
Bo-Kyeong Kim, Hyoung-Kyu Song, Thibault Castells, and Shinkook Choi.
\newblock On architectural compression of text-to-image diffusion models.
\newblock 2023{\natexlab{a}}.

\bibitem[Kim et~al.(2023{\natexlab{b}})Kim, Song, Castells, and Choi]{diffusion_prune1}
Bo-Kyeong Kim, Hyoung-Kyu Song, Thibault Castells, and Shinkook Choi.
\newblock Bk-sdm: A lightweight, fast, and cheap version of stable diffusion.
\newblock \emph{arXiv preprint arXiv:2305.15798}, 2023{\natexlab{b}}.

\bibitem[Kim et~al.(2024)Kim, Gao, Hsu, Shen, and Jin]{vit-tofo}
Minchul Kim, Shangqian Gao, Yen-Chang Hsu, Yilin Shen, and Hongxia Jin.
\newblock Token fusion: Bridging the gap between token pruning and token merging.
\newblock In \emph{Proceedings of the IEEE/CVF Winter Conference on Applications of Computer Vision}, pages 1383--1392, 2024.

\bibitem[Li et~al.(2023{\natexlab{a}})Li, Hu, Shahbaz~Khan, Li, Yang, Wang, Cheng, and Yang]{FasterDiffusion}
Senmao Li, Taihang Hu, Fahad Shahbaz~Khan, Linxuan Li, Shiqi Yang, Yaxing Wang, Ming-Ming Cheng, and Jian Yang.
\newblock Faster diffusion: Rethinking the role of unet encoder in diffusion models.
\newblock \emph{arXiv e-prints}, pages arXiv--2312, 2023{\natexlab{a}}.

\bibitem[Li et~al.(2022)Li, Thickstun, Gulrajani, Liang, and Hashimoto]{li2022diffusion}
Xiang Li, John Thickstun, Ishaan Gulrajani, Percy~S Liang, and Tatsunori~B Hashimoto.
\newblock Diffusion-lm improves controllable text generation.
\newblock \emph{Advances in Neural Information Processing Systems}, 35:\penalty0 4328--4343, 2022.

\bibitem[Li et~al.(2023{\natexlab{b}})Li, Wang, Jin, Hu, Chemerys, Fu, Wang, Tulyakov, and Ren]{diffusion_prune2}
Yanyu Li, Huan Wang, Qing Jin, Ju Hu, Pavlo Chemerys, Yun Fu, Yanzhi Wang, Sergey Tulyakov, and Jian Ren.
\newblock Snapfusion: Text-to-image diffusion model on mobile devices within two seconds.
\newblock \emph{arXiv preprint arXiv:2306.00980}, 2023{\natexlab{b}}.

\bibitem[Li et~al.(2024)Li, Wang, Jin, Hu, Chemerys, Fu, Wang, Tulyakov, and Ren]{SnapFu}
Yanyu Li, Huan Wang, Qing Jin, Ju Hu, Pavlo Chemerys, Yun Fu, Yanzhi Wang, Sergey Tulyakov, and Jian Ren.
\newblock Snapfusion: Text-to-image diffusion model on mobile devices within two seconds.
\newblock \emph{Advances in Neural Information Processing Systems}, 36, 2024.

\bibitem[Lin et~al.(2014)Lin, Maire, Belongie, Hays, Perona, Ramanan, Doll{\'a}r, and Zitnick]{coco}
Tsung-Yi Lin, Michael Maire, Serge Belongie, James Hays, Pietro Perona, Deva Ramanan, Piotr Doll{\'a}r, and C~Lawrence Zitnick.
\newblock Microsoft coco: Common objects in context.
\newblock In \emph{Computer Vision--ECCV 2014: 13th European Conference, Zurich, Switzerland, September 6-12, 2014, Proceedings, Part V 13}, pages 740--755. Springer, 2014.

\bibitem[Liu et~al.(2022{\natexlab{a}})Liu, Ren, Lin, and Zhao]{PNM}
Luping Liu, Yi Ren, Zhijie Lin, and Zhou Zhao.
\newblock Pseudo numerical methods for diffusion models on manifolds.
\newblock \emph{arXiv preprint arXiv:2202.09778}, 2022{\natexlab{a}}.

\bibitem[Liu et~al.(2022{\natexlab{b}})Liu, Ren, Lin, and Zhao]{plms}
Luping Liu, Yi Ren, Zhijie Lin, and Zhou Zhao.
\newblock Pseudo numerical methods for diffusion models on manifolds.
\newblock \emph{arXiv preprint arXiv:2202.09778}, 2022{\natexlab{b}}.

\bibitem[Luo et~al.(2023)Luo, Chen, Zhang, Huang, Wang, Shen, Zhao, Zhou, and Tan]{video-2}
Zhengxiong Luo, Dayou Chen, Yingya Zhang, Yan Huang, Liang Wang, Yujun Shen, Deli Zhao, Jingren Zhou, and Tieniu Tan.
\newblock Videofusion: Decomposed diffusion models for high-quality video generation.
\newblock In \emph{2023 IEEE/CVF Conference on Computer Vision and Pattern Recognition (CVPR)}, pages 10209--10218. IEEE, 2023.

\bibitem[Ma et~al.(2024{\natexlab{a}})Ma, Fang, Mi, and Wang]{l2cache}
Xinyin Ma, Gongfan Fang, Michael~Bi Mi, and Xinchao Wang.
\newblock Learning-to-cache: Accelerating diffusion transformer via layer caching.
\newblock \emph{arXiv preprint arXiv:2406.01733}, 2024{\natexlab{a}}.

\bibitem[Ma et~al.(2024{\natexlab{b}})Ma, Fang, and Wang]{DeepCache}
Xinyin Ma, Gongfan Fang, and Xinchao Wang.
\newblock Deepcache: Accelerating diffusion models for free.
\newblock In \emph{Proceedings of the IEEE/CVF Conference on Computer Vision and Pattern Recognition}, pages 15762--15772, 2024{\natexlab{b}}.

\bibitem[Ma et~al.(2024{\natexlab{c}})Ma, Fang, and Wang]{sd-deepcache}
Xinyin Ma, Gongfan Fang, and Xinchao Wang.
\newblock Deepcache: Accelerating diffusion models for free.
\newblock In \emph{Proceedings of the IEEE/CVF Conference on Computer Vision and Pattern Recognition}, pages 15762--15772, 2024{\natexlab{c}}.

\bibitem[Marin et~al.(2021)Marin, Chang, Ranjan, Prabhu, Rastegari, and Tuzel]{TokenPool}
Dmitrii Marin, Jen-Hao~Rick Chang, Anurag Ranjan, Anish Prabhu, Mohammad Rastegari, and Oncel Tuzel.
\newblock Token pooling in vision transformers.
\newblock \emph{arXiv preprint arXiv:2110.03860}, 2021.

\bibitem[Meng et~al.(2023{\natexlab{a}})Meng, Rombach, Gao, Kingma, Ermon, Ho, and Salimans]{DistGDM}
Chenlin Meng, Robin Rombach, Ruiqi Gao, Diederik Kingma, Stefano Ermon, Jonathan Ho, and Tim Salimans.
\newblock On distillation of guided diffusion models.
\newblock In \emph{Proceedings of the IEEE/CVF Conference on Computer Vision and Pattern Recognition}, pages 14297--14306, 2023{\natexlab{a}}.

\bibitem[Meng et~al.(2023{\natexlab{b}})Meng, Rombach, Gao, Kingma, Ermon, Ho, and Salimans]{meng2023distillation}
Chenlin Meng, Robin Rombach, Ruiqi Gao, Diederik Kingma, Stefano Ermon, Jonathan Ho, and Tim Salimans.
\newblock On distillation of guided diffusion models.
\newblock In \emph{Proceedings of the IEEE/CVF Conference on Computer Vision and Pattern Recognition}, pages 14297--14306, 2023{\natexlab{b}}.

\bibitem[Paszke et~al.(2017)Paszke, Gross, Chintala, Chanan, Yang, DeVito, Lin, Desmaison, Antiga, and Lerer]{pytorch}
Adam Paszke, Sam Gross, Soumith Chintala, Gregory Chanan, Edward Yang, Zachary DeVito, Zeming Lin, Alban Desmaison, Luca Antiga, and Adam Lerer.
\newblock Automatic differentiation in pytorch.
\newblock 2017.

\bibitem[Rao et~al.(2021)Rao, Zhao, Liu, Lu, Zhou, and Hsieh]{Dynavit}
Yongming Rao, Wenliang Zhao, Benlin Liu, Jiwen Lu, Jie Zhou, and Cho-Jui Hsieh.
\newblock Dynamicvit: Efficient vision transformers with dynamic token sparsification.
\newblock \emph{Advances in neural information processing systems}, 34:\penalty0 13937--13949, 2021.

\bibitem[Rombach et~al.(2022)Rombach, Blattmann, Lorenz, Esser, and Ommer]{stable-diffusion}
Robin Rombach, Andreas Blattmann, Dominik Lorenz, Patrick Esser, and Bj{\"o}rn Ommer.
\newblock High-resolution image synthesis with latent diffusion models.
\newblock In \emph{Proceedings of the IEEE/CVF conference on computer vision and pattern recognition}, pages 10684--10695, 2022.

\bibitem[Ronneberger et~al.(2015)Ronneberger, Fischer, and Brox]{unet}
Olaf Ronneberger, Philipp Fischer, and Thomas Brox.
\newblock U-net: Convolutional networks for biomedical image segmentation.
\newblock In \emph{Medical image computing and computer-assisted intervention--MICCAI 2015: 18th international conference, Munich, Germany, October 5-9, 2015, proceedings, part III 18}, pages 234--241. Springer, 2015.

\bibitem[Salimans and Ho(2022)]{pd}
Tim Salimans and Jonathan Ho.
\newblock Progressive distillation for fast sampling of diffusion models.
\newblock \emph{arXiv preprint arXiv:2202.00512}, 2022.

\bibitem[Salimans et~al.(2016)Salimans, Goodfellow, Zaremba, Cheung, Radford, and Chen]{isscore}
Tim Salimans, Ian Goodfellow, Wojciech Zaremba, Vicki Cheung, Alec Radford, and Xi Chen.
\newblock Improved techniques for training gans.
\newblock \emph{Advances in neural information processing systems}, 29, 2016.

\bibitem[Shang et~al.(2023{\natexlab{a}})Shang, Yuan, Xie, Wu, and Yan]{diffusion_quant1}
Yuzhang Shang, Zhihang Yuan, Bin Xie, Bingzhe Wu, and Yan Yan.
\newblock Post-training quantization on diffusion models.
\newblock In \emph{Proceedings of the IEEE/CVF Conference on Computer Vision and Pattern Recognition}, pages 1972--1981, 2023{\natexlab{a}}.

\bibitem[Shang et~al.(2023{\natexlab{b}})Shang, Yuan, Xie, Wu, and Yan]{diffusion_quant2}
Yuzhang Shang, Zhihang Yuan, Bin Xie, Bingzhe Wu, and Yan Yan.
\newblock Post-training quantization on diffusion models.
\newblock In \emph{Proceedings of the IEEE/CVF conference on computer vision and pattern recognition}, pages 1972--1981, 2023{\natexlab{b}}.

\bibitem[Sohl-Dickstein et~al.(2015)Sohl-Dickstein, Weiss, Maheswaranathan, and Ganguli]{dm-1}
Jascha Sohl-Dickstein, Eric Weiss, Niru Maheswaranathan, and Surya Ganguli.
\newblock Deep unsupervised learning using nonequilibrium thermodynamics.
\newblock In \emph{International conference on machine learning}, pages 2256--2265. PMLR, 2015.

\bibitem[Song et~al.(2020)Song, Meng, and Ermon]{ddim}
Jiaming Song, Chenlin Meng, and Stefano Ermon.
\newblock Denoising diffusion implicit models.
\newblock \emph{arXiv preprint arXiv:2010.02502}, 2020.

\bibitem[Song and Ermon(2019)]{dm-2}
Yang Song and Stefano Ermon.
\newblock Generative modeling by estimating gradients of the data distribution.
\newblock \emph{Advances in neural information processing systems}, 32, 2019.

\bibitem[Song et~al.(2023)Song, Dhariwal, Chen, and Sutskever]{CM}
Yang Song, Prafulla Dhariwal, Mark Chen, and Ilya Sutskever.
\newblock Consistency models.
\newblock \emph{arXiv preprint arXiv:2303.01469}, 2023.

\bibitem[Wang et~al.(2024)Wang, Liu, Kang, Li, Lin, Jha, and Liu]{sd-adt}
Hongjie Wang, Difan Liu, Yan Kang, Yijun Li, Zhe Lin, Niraj~K Jha, and Yuchen Liu.
\newblock Attention-driven training-free efficiency enhancement of diffusion models.
\newblock In \emph{Proceedings of the IEEE/CVF Conference on Computer Vision and Pattern Recognition}, pages 16080--16089, 2024.

\bibitem[Yin et~al.(2022)Yin, Vahdat, Alvarez, Mallya, Kautz, and Molchanov]{A-vit}
Hongxu Yin, Arash Vahdat, Jose~M Alvarez, Arun Mallya, Jan Kautz, and Pavlo Molchanov.
\newblock A-vit: Adaptive tokens for efficient vision transformer.
\newblock In \emph{Proceedings of the IEEE/CVF conference on computer vision and pattern recognition}, pages 10809--10818, 2022.

\bibitem[Zhang et~al.(2023)Zhang, Dong, Tai, and Ma]{pointdistiller}
Linfeng Zhang, Runpei Dong, Hung-Shuo Tai, and Kaisheng Ma.
\newblock Pointdistiller: structured knowledge distillation towards efficient and compact 3d detection.
\newblock In \emph{Proceedings of the IEEE/CVF conference on computer vision and pattern recognition}, pages 21791--21801, 2023.

\bibitem[Zhao et~al.(2024)Zhao, Ning, Fang, Liu, Huang, Lin, Yan, Dai, and Wang]{diffusion_quant3}
Tianchen Zhao, Xuefei Ning, Tongcheng Fang, Enshu Liu, Guyue Huang, Zinan Lin, Shengen Yan, Guohao Dai, and Yu Wang.
\newblock Mixdq: Memory-efficient few-step text-to-image diffusion models with metric-decoupled mixed precision quantization.
\newblock \emph{arXiv preprint arXiv:2405.17873}, 2024.

\end{thebibliography}
}


\end{document}